\documentclass[conference]{IEEEconf}

\input epsf

\usepackage{titlesec}
\usepackage{balance} 
\usepackage{stfloats}
\usepackage{xcolor}
\RequirePackage[most]{tcolorbox}

\usepackage{listings}
\usepackage{siunitx}
\usepackage{multirow}

\usepackage{amsmath,amssymb,amsfonts}
\usepackage{algorithmic}
\usepackage{graphicx}
\graphicspath{{images/}}
\usepackage{textcomp}

\usepackage{url}
\usepackage{caption}
\DeclareCaptionLabelFormat{nospace}{#1#2}
\renewcommand{\thetable}{\arabic{table}}

\renewcommand\thesection{\arabic{section}} 

\renewcommand\thesubsectiondis{\thesection.\arabic{subsection}}

\renewcommand\thesubsubsectiondis{\thesubsectiondis.\arabic{subsubsection}}

\begin{document}
\title{\textbf{\Large Log-based Anomaly Detection of Enterprise Software: An Empirical Study\\}}

\author{Nadun Wijesinghe$^{1}$, Hadi Hemmati$^{2,*}$\\
	\normalsize $^{1}$University of Calgary, Calgary, AB, Canada\\
	\normalsize $^{2}$York University, Toronto, ON, Canada\\
	\normalsize npwijesi@ucalgary.ca, hemmati@yorku.ca\\
	\normalsize *corresponding author
}


\maketitle

\begin{abstract}
Most enterprise applications use logging as a mechanism to diagnose anomalies, which could help with reducing system downtime. Anomaly detection using software execution logs has been explored in several prior studies, using both classical and deep neural network-based machine learning models. In recent years, the research has largely focused in using variations of sequence-based deep neural networks (e.g., Long-Short Term Memory and Transformer-based models) for log-based anomaly detection on open-source data. However, they have not been applied in industrial datasets, as often. In addition, the studied open-source datasets are typically very large in size with logging statements that do not change much over time, which may not be the case with a dataset from an industrial service that is relatively new. In this paper, we evaluate several state-of-the-art anomaly detection models on an industrial dataset from our research partner, which is much smaller and loosely structured than most large scale open-source benchmark datasets. Results show that while all models are capable of detecting anomalies, certain models are better suited for less-structured datasets. We also see that model effectiveness changes when a common data leak associated with a random train-test split in some prior work is removed. A qualitative study of the defects' characteristics identified by the developers on the industrial dataset further shows strengths and weaknesses of the models in detecting different types of anomalies. Finally, we explore the effect of limited training data by gradually increasing the training set size, to evaluate if the model effectiveness does depend on the training set size.
\end{abstract}

\IEEEoverridecommandlockouts
\vspace{1.5ex}

\section{Introduction}
Software in the present day consists of many interconnected systems that rely on each other to handle requests. Recording run-time logs is a common practice in such systems~\cite{zhu2015}, and the logs are frequently the main source for debugging~\cite{archen2019}. Logs are routinely written by an application to a central location, often at various logging priority levels (e.g., error, warning, info etc.). As logging is done in real-time, logs provide an insight into the current system health and how it is failed. This then points to the natural next step: by monitoring the system logs in real-time, we can quickly and accurately detect when system failures occur, reducing system downtime.

Large-scale industrial applications often have a large amount of logs being printed every second~\cite{zhang2019}, making manual inspection of those logs in real-time a difficult task. This is further compounded by different types of errors, and errors that require inspecting the entire event sequence, as opposed to a single log line. Servers also handle multiple requests in parallel, resulting in data from different requests being logged at the same time. All those issues, combined with the sheer volume of logs printed by large-scale complex applications, have rendered real-time manual inspection unfeasible. Therefore, an automated approach is required to shift through all the data and correctly detect anomalies.

A common automated method used in the industry to detect anomalies using logs is to utilize deterministic rules. By creating a set of regular expressions or search terms, it's possible to search through the log sequences to find potentially anomalous events. The search terms could include words such as ``error'' or ``failure''. While this does help with detecting the more obvious of anomalies, it is not accurate: those search terms could easily appear in non-anomalous sequence, or it could be an error that is self-corrected and does not result in a more severe anomaly. In addition, some anomalies could be a result of an incorrect sequence, which may appear benevolent on the surface. For example, if a file is opened, but not closed, the log sequence may not show any errors: it would just end before the closing of the file. Those anomalies cannot be detected by the use of a deterministic algorithm. In order to evaluate whether the sequence is anomalous or not, the entire sequence needs to be considered, not just the current log event.

To evaluate its effects, this type of deterministic algorithm was implemented on our industrial partner's logs, and the results were less than satisfactory. Results in terms of F1-score, one of the metrics that will be defined in section \ref{sc:metrics}, were 40\% below that of the lowest performing state-of-the-art anomaly detection model.

Therefore, in this paper, our goal is to study the effectiveness of state of the art log anomaly detection techniques. Our data comes from our industrial partner's microservice hosted on AWS Lambda, whose logs are collected from AWS CloudWatch. 

There have been several studies done on evaluating automatic detection of anomalies using execution logs in the past, but most these studies are on large open-source data. We identify some differences between the characteristics of such studies and our interest in this paper (industrial system with limited logs), as follows:
\begin{enumerate}
  \item \textbf{Data size}: Open-source log collections have large amounts of data, usually over 10 million lines of logs \cite{logpai}. A recently developed microservice in a small company, for example, may only have logs in the range of thousands, often produced by the underlying platform during testing. If this microservice is deemed sensitive to the business, the anomaly detection model will need to be trained (or fine-tuned) and evaluated on the limited dataset, before being deployed to production. Therefore, the evaluation mechanism of the models stated above using large open-source datasets and the corresponding findings may not hold during many industrial practices.
  \item \textbf{Data uniformity}: As most of the open-source log collections are typically retrieved from well-established industrial applications, they are often uniform in nature. This means the logs do not change much over time, due to modifications of source code logging statements. In a relatively new industrial application, the source code logs would be updated frequently to show missing data, and would undergo additions/removals of logging statements as well. In addition, the logs may lack structure, and have more of a natural language free format of logging information. These aspects are less studied in most of the automated log anomaly detection methods.
\end{enumerate}

To fill this gap (evaluation of log anomaly detection for small industrial datasets), in this paper, we have selected several state of the art models and trained them on our datasets.


Replication package: The source code for analyzing data is publicly available at \url{https://zenodo.org/record/7553290}. This includes the open-source dataset as well. Please note that the industrial dataset has not been included due to confidentiality reasons.

\section{Background}
\label{bkg}
Log anomaly detection is generally performed in 3 steps: Pre-processing, Model training and Prediction. As logs are written by developers, they are often in free-form natural language format, and the pre-processing step includes converting them into a more structured form, which can then be fed into a model for training. Common pre-processing steps include:

\begin{enumerate}
  \item \textbf{Log cleaning}: A single log line often comprises of several elements, such as timestamp, log level and log content. During this step, the log content is extracted and added as a property of an object. This object may also contain the log level and timestamp, based on the model requirements.
  \item \textbf{Template mining}: A log comprises of two distinct parts: a constant and a variable. The constant part, often also called \textit{log template} or \textit{log key}, has the overall structure of the log statement, as well as words that do not change for each statement. The variable part includes the parameters of the log, which may change for each log statement of its kind.  For example, a log message such as \textit{``Received block 3587508140051953248 of size 67108864"} can be split into the log key \textit{``Received block * of size *"}, and the parameter values \textit{[3587508140051953248, 67108864]}. Popular log template miners include Spell~\cite{spell}, which uses a longest common sub-sequence based approach, and Drain~\cite{drain}, which uses a parse tree with a fixed depth during the log group search.
  \item \textbf{Sequence generation}: Since multiple requests can be handled by a system at a given moment, log messages corresponding to a specific request would be scattered among other log statements. During this step, logs corresponding to each request/event is grouped together, by using a field that denotes the request/event that log belongs to. The field name differs between systems: Hadoop Distributed File System (HDFS) for instance uses \textit{block ID}, while AWS CloudWatch uses \textit{request ID}.
  \item \textbf{Vectorization (optional)}: While log sequences can be directly used to train a model, several papers go a step further to vectorize the log keys. During this step, each word of the log key is converted to a vector (by using a method such as Word2Vec) and then aggregated over the entire log. Additional steps would include semantic information integration, such as replacement of synonyms by using a lexical database and domain knowledge. The final result is either a one-dimensional or 2-dimensional array for each log key.
\end{enumerate}

The type of model varies between papers, but can be divided into 2 broad categories: classical techniques and deep learning techniques.

\subsection{Classical Techniques}

Classical techniques for anomaly detection have been in use for over a decade. Statistical methods, such as statistical workflow execution~\cite{chen2002}, state machine-based modeling~\cite{tan2008} and Hidden Markov Models~\cite{yamanishi2005} have been used in several papers. Frequent pattern mining, also called invariant mining, has also been used with promising results~\cite{lou2010}~\cite{xu20092}~\cite{farshchi2015}. 

Classical machine learning based approaches have also been used in several papers, with varying degrees of success. Classical machine learning based approaches include modeling anomaly detection as a type of machine learning algorithm. One possible method is to treat this as a clustering problem, where normal log events need to be in one cluster and anomaly events should be far away from the normal cluster (which may themselves belong to either a single cluster or multiple clusters). LogCluster used a similar mechanism, and used the centroid of each cluster to depict the log sequence~\cite{lin2016}. This log sequence could then be used to identify the underlying root cause. Log3C used a method called Cascading Clustering to group log sequences, and a linear regression model to find issues related to deterioration of system KPIs~\cite{he2018}. Another approach is to use classification methods to categorize input log sequences into normal or anomalous sequences, using supervised learning. Logistic Regression was used in one paper on several open-source datasets, using event count vectors~\cite{he2016}. Liang~\cite{liang2007} used a Support Vector Machine (SVM) based model and a K-Nearest Neighbour (KNN) model on vectorized logs, using several features such as event counts. Another paper proposed an SVM based model that used empirical properties of logs, such as frequency and periodicity, to detect anomalies~\cite{kimura2015}.

A different approach was taken in another experiment, where anomaly detection was modeled as a dimensionality reduction problem~\cite{xu2009}. This was done by transforming the data to rely on limited dimensions (as opposed to the high dimensionality of the original dataset). Anomaly detection was then performed by identifying logs at a distance higher than a specified threshold. An important aspect to note here is that they used parameter value vectors in addition to log event count, as parameter value vectors are often unused in machine learning models. Graphical methods have also been used in this area of research, with promising results. A CFG mining method was used to detect sequence and distribution anomalies on synthetic traces and log datasets~\cite{nandi2016}. Another paper employed statistical inference methods to infer dependencies among log events, by using a 3 step algorithm~\cite{lou20102}. NLP based log parsing methods were evaluated in a different paper, by using different n-gram models and hashing~\cite{aussel2018}. Modelling was done via Bisecting K-Means and Latent Dirichlet Allocation (LDA), followed by Random Forest for classification.

While classical methods have shown some performance, they have recently been out-shined by models based on deep neural networks.

\subsection{Deep Learning Techniques}

In deep learning techniques, the log templates/vectors of the training set are fed into a neural network, and validated with a separate set of data. Most models use Long Short-Term Memory (LSTM), which is a type of Recurrent Neural Network (RNN), with promising results.

\subsubsection{LSTM Based Techniques}

LogRobust is one such model~\cite{zhang2019}. LogRobust used Drain~\cite{drain} to mine for log templates, followed by a textual preprocessing stage to assist with word vectorization. The words were then vectorized using FastText~\cite{fasttext}, then aggregated using TF-IDF of each word, so that a single vector represented a single log event. The goal behind semantic aggregation was to reduce noise in the data, which can occur from incorrect log template parsing and continuously evolving logs. The TF-IDF aggregation ensured that the effect of evolving log statements and incorrect log parsing was reduced, resulting in a more uniform vector for each variation of log event. An attention-based bidirectional LSTM was then trained using the semantic vectors, with injected instabilities in the logs. The model results were compared to classical methods, and showed promising results. Vinayakumar~\cite{vinayakumar2017} proposed a similar model, which used a stacked-LSTM (created by adding recurrent LSTM layers on top of existing LSTM layers) and was trained with both normal and anomalous logs. Experiments were then performed to optimize the hyper-parameters of the model using the CDMC2016 dataset. Wang~\cite{wang2018} compared performances of different natural language processing feature extraction methods, namely Word2Vec and TF-IDF. The extracted features were fed into an LSTM to perform anomaly detection, and compared against Gradient Boosting Decision Tree (GBDT) and Naive Bayes methods. Zhao~\cite{zhao2021} proposed a tokenization method based on ASCII values of log characters. Each log event was considered a sentence, and converted to a string of ASCII values normalized to start with 0. This was then used to train an LSTM model. Another paper explored the temporal-spatial information in microservices, in the form of logs (temporal) and query traces (spatial)~\cite{zuo2020}. By combining the data and training an LSTM model, they were able to segregate anomalies from normal data and detect failures.

Several other models in this category are evaluated on this paper (DeepLog\cite{deeplog}, LogAnomaly\cite{loganomaly} and LogBERT\cite{logbert}), and are explained in detail in the next section. They were selected due to their high metrics, recentness and readily-available source code.

\subsubsection{Other Techniques}

While RNNs have been most widely-used type of neural network, there have been some research on other types of neural networks as well. Liu~\cite{liu2019} proposed a model based on Gated Recurrent Unit (GRU) networks, which is another type of RNN, combined with a Support Vector Data Description (SVDD) model. PCA was initially applied on the dataset to reduce dimensionality, followed by the GRU-SVDD model. This was evaluated on classical KDD Cup99 datasets. Lu~\cite{lu2018} used a Convolutional Neural Network (CNN) for anomaly detection. After parsing log events into log templates, a custom trainable matrix called Logkey2Vec was used to map each log key into a vector. The CNN was then trained using the vectorized logs, from the HDFS dataset. Another model, LogGAN, was proposed as an LSTM-based Generative Adversarial Network (GAN)~\cite{xia2021}. By using a generator and a discriminator, LogGAN was able to analyze distribution of the training set and create artificial data points. This was then used to mitigate data imbalance between normal and anomalous datasets.

Failure prediction and diagnosis are two other avenues of log-based reliability engineering, but are out of scope for this paper, and therefore have not been explored in detail.

Among these papers, only a limited set have applied models on industrial datasets. To our knowledge, this is the first time those state-of-the-art models (DeepLog, LogAnomaly and LogBERT) have been evaluated in an industrial dataset, with limitations on data size and uniformity.

\section{Methodology}

\subsection{Objectives and RQs}

The main objective of this study is to evaluate the effectiveness of current state-of-the-art anomaly detection models on an industrial dataset, to determine which would be the best candidate to be potentially deployed to production. To address the this objective, we explore the following research questions in this paper:

\newcommand{\RQOne}{How effective are the current log anomaly detection models in detecting failures, in an industrial dataset?}

\newcommand{\RQTwo}{Does the type of train-test splitting affect the reported effectiveness of the models?}

\newcommand{\RQThree}{How successful are the models in detecting different types of failures?}

\newcommand{\RQFour}{How does the size of training set affect the model effectiveness?}

\noindent \textbf{RQ1:} \RQOne\ \\
The goal of this research question is to explore the effectiveness of each model applied to an industrial dataset. We compare and contrast our findings with reported results on a well-known open source dataset in this domain (the HDFS dataset). \\
\noindent \textbf{RQ2:} \RQTwo\ \\
There has been a recent observation in the literature of a data leak during the random sampling process~\cite{hoang2022} in many similar studies. The issue is that using a random split for generating train/test sets results in future logs potentially being used to predict past logs, resulting in incorrect reports of metrics. In this research question we explore the effect of using a time-based split instead of a random split on both the industrial dataset and HDFS dataset. \\
\noindent \textbf{RQ3:} \RQThree\ \\
In RQ3, we conduct a qualitative analysis of different types of anomalies in the industrial dataset. The goal is to explore if the failures can be categorized to types (identified by our industrial partner), and to see if certain models are better at predicting certain types of failures. \\
\noindent \textbf{RQ4:} \RQFour\ \\
One of the driving factors of this research is the impact of limited data on training anomaly detection models. In this research question, we aim to find if the model effectiveness can be improved by increasing the training set size. \\

\subsection{Datasets}

\begin{figure}[!tbp]
    \centering
    \includegraphics[width=0.45\textwidth]{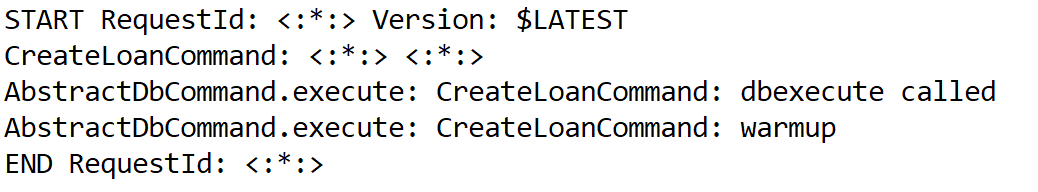}
    \caption{Sample logs from the industrial microservice}
    \label{fig:awslogs}
\end{figure}

\begin{figure}[!tbp]
    \centering
    \includegraphics[width=0.45\textwidth]{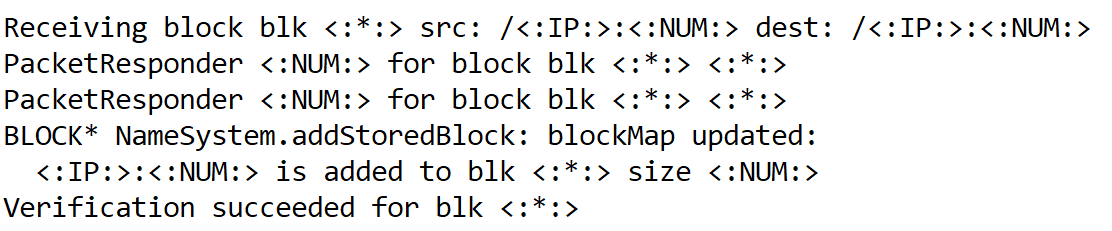}
    \caption{Sample logs from the open-source HDFS dataset}
    \label{fig:hdfslogs}
\end{figure}

The models are evaluated on two datasets:
\begin{enumerate}
  \item \textbf{Industrial dataset}: A collection of logs from an industrial application. This application is a microservice hosted on AWS Lambda, which deals with incoming requests from a user application. The logs themselves have been collected from AWS CloudWatch. Both CloudWatch and Lambda are part of the Amazon Web Services cloud platform~\cite{aws}. As part of the larger AWS features, CloudWatch has built-in support for log correlation, visualization and monitoring. A set of sample log templates can be found in figure \ref{fig:awslogs} (raw logs have not been shown here due to confidentiality reasons).
  \item \textbf{Open-source dataset}: the HDFS dataset, from the Logpai repository~\cite{logpai}. The HDFS dataset has been widely used to benchmark anomaly detection models, and comes with over 10 million lines of logs that have been labeled. A selected set of log templates from this dataset can be found in figure \ref{fig:hdfslogs}.
\end{enumerate}

The industrial microservice is a small-to-medium scale user-facing application, and is assumed to be dealing with approximately a hundred user requests per day in production. The endpoint itself has been in use for over 2 years, and has recently undergone changes to allow for containerization. Due to this, it was decided to retrieve the more recent logs, in order to ensure stale data are not used to train the models.

Anomalies on this application are not monitored. Therefore, the time to detect an anomaly could range between several minutes or hours, depending on the severity of the anomaly (which could raise alarms in a related but different application) and based on how often a developer may manually check the logs. In the event of an anomaly, the system could be unresponsive for some time, or it could be returning failure responses to the systems that call it. Those anomalies may require restarting the service, or deploying a code change to fix the underlying issue. Therefore, an anomaly detection model could have a significant impact on reducing the application downtime.

Table \ref{tab:dataset_properties} outlines the properties of the two datasets. As can be seen from the table, the industrial dataset is much smaller than the HDFS dataset, which is in-line with the characteristic mentioned earlier regarding dataset size. The template count, on the other hand, is much higher in the industrial dataset, even with its small size. This shows that it is loosely-structured compared to the HDFS dataset, with the logs appearing in a more natural language free-form format. The time duration is also higher in the industrial dataset, which shows that as a new microservice, a longer time duration is required to gather logs (since it does not get requests as often).

\begin{table}[!tbp]
    \caption{Dataset properties \protect\label{tab:dataset_properties} }
    \begin{center}
      \begin{tabular}{|p{1.5cm}|p{1.5cm}|p{1.5cm}|p{1.5cm}|}
        \hline
        \textbf{Dataset} & \textbf{Number of logs} & \textbf{Templates count} & \textbf{Time \slash duration}\\
        \hline
        Industrial microservice & 170,566 & 142 & 2 months\\
        HDFS & 11,175,629 & 53 & 39 hours\\
        \hline
        \end{tabular}
    \end{center}
\end{table}

\subsection{Models}

For all models, the logs undergo a preprocessing stage, where data such as timestamps and large objects are removed from the logs. Afterwards, Drain3~\cite{drain} is used to mine for log templates (as explained in the Section \ref{bkg}), and log sequences are generated for each sequence ID. Then the logs are split into 3 parts: training set, normal test set and anomaly test set. Then a sliding window method is used to generate history sequences. After training the model with the given training set, the two test sets are used to evaluate model performance.

It is important to note that the training set constitutes of only \textbf{normal logs}. This is due to the fact that anomalous logs are rare to find in real-world, and there is a high data skew towards normal data points in any industrial dataset. This is followed in all the models evaluated in this study.

\begin{figure*}
    \centering
    \includegraphics[width=0.95\textwidth]{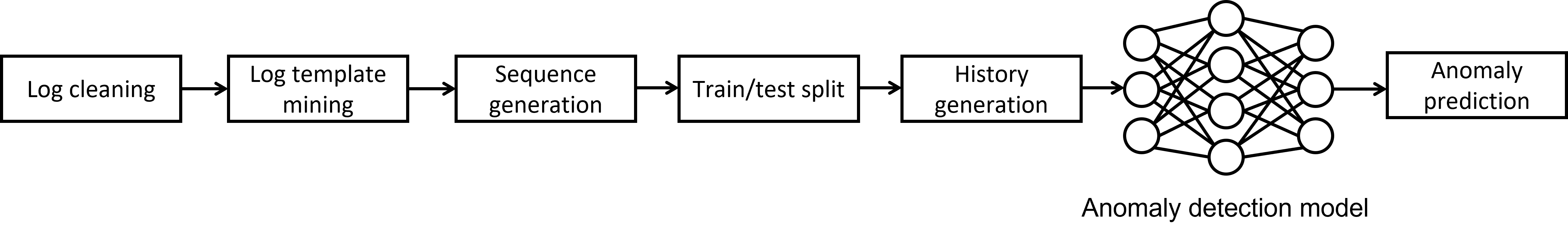}
    \caption{High level illustration of the log anomaly detection process.}
    \label{fig:experiment_flow}
\end{figure*}

A diagram of the experiment workflow can be found in figure \ref{fig:experiment_flow}.

In this paper, we compare the following state-of-the-art models published in past papers:
\begin{itemize}
  \item DeepLog~\cite{deeplog}
  \item LogAnomaly~\cite{loganomaly}
  \item LogBERT~\cite{logbert}
  \item Baseline LSTM Model
\end{itemize}

These models are selected because (1) they have showed great results in their original paper, (2) they are relatively recent (all have been published in the last 6 years), and (3) their source-code is readily available and is replicable. In the following sub-section, we explain them in details.

\subsubsection{DeepLog}

DeepLog~\cite{deeplog} is an LSTM model for anomaly detection, but unlike most models, DeepLog considers both the log template as well as the parameter values of the log event as inputs. DeepLog consists of two main models: a log key anomaly detection model and a parameter value anomaly detection model. The log key sequence is used to train the log key anomaly detection model, which uses an LSTM. This is treated as a multi-class classification problem, where each distinct log key is assigned a class. By using the history of recent log keys (named \textit{window size}), the model outputs the probability distribution over all log keys for the next log. If the actual log is included in the top \textit{k} candidates, it is treated as non-anomalous. If it is not included, then it is classified as an anomaly. The parameter value anomaly detection model uses a separate LSTM, and is trained with parameter vectors, which includes variable parts of the log as well as the time difference between the current and previous log event. The model outputs a predicted parameter value vector, which is then compared with the one from the actual log event. Error is calculated as the difference between predicted and actual, and compared against previously generated Gaussian error distribution. If it is within a pre-specified confidence interval, it is classified as normal, and as an anomaly otherwise.

In addition to anomaly detection, DeepLog also includes a workflow creation aspect. By using the anomaly detection model and a density clustering approach, DeepLog can re-create the flow that resulted in a failure, allowing developers to investigate root causes of failures more effectively. While this is not part of anomaly detection itself, it provides a significant enhancement for failure diagnosis.

\subsubsection{LogAnomaly}

LogAnomaly~\cite{loganomaly} is another model proposed within the last few years for anomaly detection. It considers anomalies of 2 types: sequential anomalies and quantitative anomalies. Sequential anomalies occur when a log sequence deviates from normal patterns, which are learned during the training phase as training data only consists of normal logs. This is a common method for anomaly detection, which is also used by several other models described earlier. Where LogAnomaly differs, however, is when it comes to the second anomaly type: Quantitative anomalies occur when linear relationships are broken between log sequences. For example, a normal sequence may have one log statement for opening a file and another for closing it. This is a 1:1 relationship between two logs, which means the number of file opening logs should equal the number of file closing ones. In a test sequence, if a file opens but does not close, this would constitute an anomaly, and would be detected as the linear relationship between the two logs has now been violated.

LogAnomaly starts off by using Frequent Template Tree (FT-Tree)~\cite{zhang2017} to parse logs and extract templates. The log templates are then encoded by using a mechanism named \textit{Template2Vec}. In \textit{Template2Vec}, first a set of synonyms and antonyms is created using the lexical database WordNet and domain knowledge. Then the distributed lexical-contrast embedding (dLCE) model is used to create word vectors. Finally, template vectors (for each log template) are calculated by taking the weighted average of each word vector in the log template.

Once the template vectors have been calculated, they are fed into an LSTM model for training, which is then used to detect sequential anomalies. For quantitative anomaly detection, the model counts the different templates present in the history. During detection phase, the model predicts the next log based on its sequential history, and patterns learned during quantitative relationships training phase. This outputs a vector of probabilities for the next log, with one probability for each log template. If the actual log is observed in the top \textit{k} candidates, it is classified as normal. Otherwise, it is flagged as an anomaly.

LogAnomaly also does template approximation on new logs unseen during the training phase. This is done by extracting a temporary template using FT-Tree, calculating a template vector and matching it to an existing one based on similarity. The reasoning is that the majority of "new" templates are minor variants of existing ones, which have occurred due to small updates to a logging statement.

\subsubsection{LogBERT}

A more recent deep learning model is LogBERT~\cite{logbert}. This uses a transformer, based on BERT (Bidirectional Encoder Representations from Transformers). During preprocessing, log templates are first mined from log sequences. The templates sequence is then represented as a summation of a log key embedding (a randomly generated matrix) and position embedding (generated using a sinusoidal function). Then a transformer encoder with multiple layers of transformers is used to learn relationships in a log template sequence.

The model itself is trained by using two self-supervised training tasks: Masked Log Key Prediction (MLKP) and Volume of Hypersphere Minimization (VHM). In MLKP, LogBERT is trained to predict several masked log keys in a sequence. This is done by replacing a set of random log keys in a sequence with a \textit{[MASK]} token, and having the model predict the masked log keys. This gives the model contextual knowledge of log sequences. The underlying goal is that if normal and anomalous log sequences are sufficiently different, the contextual knowledge can be used to differentiate anomalous from normal sequences, since the model is only trained on normal logs. In VHM, a hypersphere enclosing normal data is created. Similar to Deep SVDD, the goal is to minimize the volume of this hypersphere. The motivation is that normal logs should be similar to each other in the embedding space, while anomalous should be as far from the hypersphere as possible. The final objective function is the addition of the two training tasks, with a hyper-parameter $\alpha$ that adjusts the weight of the VHM output.

The model's final output is a vector of probabilities for the next log key, similar to DeepLog and LogAnomaly above. This is of length \textit{g}, which includes the candidates with the highest probabilities. If the actual log key is in this vector, the log key is considered normal, and as an anomaly otherwise. If a log key sequence consists of more than \textit{r} log keys, the sequence itself is considered anomalous.

\subsubsection{Baseline LSTM Model}

To establish a baseline, we use a basic LSTM model. This model, similar to the models above, only use normal logs for training. The LSTM model itself is built using 4 LSTM layers, followed by a dropout layer composed of 100 neurons each. The last layer is a fully connected Dense layer with a size of \textit{X} neurons, where \textit{X} corresponds to the number of unique log templates. The last layer is a softmax, which outputs a list of probabilities for the next log, showing the most likely candidates. Anomaly detection is done by checking whether the next log is in the top \textit{k} candidates. If it is within the candidates list, it is classified as a normal sequence, otherwise, as an anomalous sequence.

The anomaly detection mechanism is identical to the state-of-the-art models, which is intentional. This allows us to test whether the additional logic parts implemented in the state-of-the-art models are useful at giving better predictions compared to using a pure LSTM model.

The model was trained using the Adam optimizer, using categorical cross-entropy as the loss function. The model hyper-parameters were calculated using a simple grid-search approach. The model details can be found in table \ref{tab:baseline_model}.

\begin{table}[!tbp]
    \caption{Baseline LSTM model parameters \protect\label{tab:baseline_model} }
    \begin{center}
      \begin{tabular}{|l|l|l|}
        \hline
        \textbf{Layer type } & \textbf{Output shape} & \textbf{Parameters count}\\
        \hline
        LSTM & (3, 100) & 40800\\
        Dropout & (3, 100) & 0\\
        LSTM & (3, 100) & 80400\\
        Dropout & (3, 100) & 0\\
        LSTM & (3, 100) & 80400\\
        Dropout & (3, 100) & 0\\
        LSTM & (3, 100) & 80400\\
        Dropout & (3, 100) & 0\\
        Dense & (170) & 17170\\
        Dense & (170) & 29070\\
        \hline
      \end{tabular}
    \end{center}
  \end{table} 

\subsection{Evaluation Metrics}
\label{sc:metrics}

Since anomaly detection is considered to be a type of classification problem, we use F1 score, precision, and recall to evaluate model performance. All values are calculated using Confusion matrix (true positives count, true negatives count, false positives count and false negatives count). We excluded the accuracy metric, which is calculated as the proportion of correct predictions out of all predictions for our anomaly detection models, due to the skew of data. Most of the logs in any dataset are normal logs, with anomaly logs being only a fraction of it. Therefore, a model can predict all logs as normal, and still end up with a high degree of accuracy.

\textbf{Precision} calculates the relevancy of the predicted results. It's calculated as the proportion of true positives out of all the positives detected. A high degree of precision is an indicator of a stable model that does not raise many unnecessary alarms.

\textbf{Recall} is calculated as the proportion of detected anomalies out of all anomalies. A model with high recall means it does not miss many anomalies.

\textbf{F1-Score} is the harmonic mean of precision and recall. The label F1 comes from the fact that both precision and recall are weighted evenly. This is often used in anomaly detection models as the single criterion to evaluate the model.

In this paper, we will be using the F1-score as the primary metric for evaluating the models. Recall will be used as the secondary metric, since it shows the ability of the model to detect actual anomalies.

\begin{eqnarray}
Precision = \frac{TP}{TP+FP}
\end{eqnarray}

\begin{eqnarray}
Recall = \frac{TP}{TP+FN}
\end{eqnarray}

\begin{eqnarray}
F1-Score = 2*\frac{Precision * Recall}{Precision + Recall}
\end{eqnarray}

\subsection{Experiment Setup}

AWS SageMaker was used as the machine-learning platform to conduct the experiments. AWS SageMaker is a cloud-based machine learning platform, that allows the creation, testing and deployment of ML models~\cite{sagemaker}. It also facilitates connectivity between other AWS services, such as CloudWatch. The industrial dataset was imported from AWS CloudWatch, and the open-source HDFS dataset was downloaded from their GitHub repository.

All experiments were run on a high-performance AWS SageMaker Notebook Instance with a high virtual CPU count and RAM, using Jupyter Notebook. Details regarding the SageMaker Notebook Instance can be found in Table \ref{tab:sagemaker_properties}.

\begin{table}[!tbp]
    \caption{AWS SageMaker Notebook Instance properties
\protect\label{tab:sagemaker_properties} }
    \begin{center}
      \begin{tabular}{|l|c|}
        \hline
        \textbf{Parameter name} & \textbf{Value}\\
        \hline
        Instance type & ml.m5.4xlarge\\
        Platform & Amazon Linux 2\\
        vCPU & 16\\
        Memory & 64 GB\\
        \hline
      \end{tabular}
    \end{center}
  \end{table}

\section{Empirical Study}

\subsection{RQs design and results}
\subsubsection{RQ1: \RQOne\ }

\paragraph{Design}

To answer this research question, all four selected models were trained on both the industrial dataset as well as the HDFS dataset. As the state-of-the-art models all used random sampling in their original papers, we used the same split type for consistency. In the original papers of the selected models, a relatively small portion is used to train the model (in the case of DeepLog, 4855 logs were used for training, which is less than 1\% of the dataset). In order to keep a valid amount of logs for testing, we opted for K-fold validation with 5 folds, with one fold used for training and the rest for testing. The results were then calculated as the mean of all the folds. Our focus in this paper is to examine the effect of having a small dataset for training, which could be even smaller than the amount of data we had for the industrial service. Therefore, using only one fold for training would give us a better idea of how each of the models perform when there is not an abundance of data.

For consistency, the training size was kept the same with the HDFS dataset. As the HDFS dataset is quite large, this allowed us to use a significant amount of it for testing.

\paragraph{Results}

\begin{figure}[!tbp]
    \centering
    \includegraphics[width=0.45\textwidth]{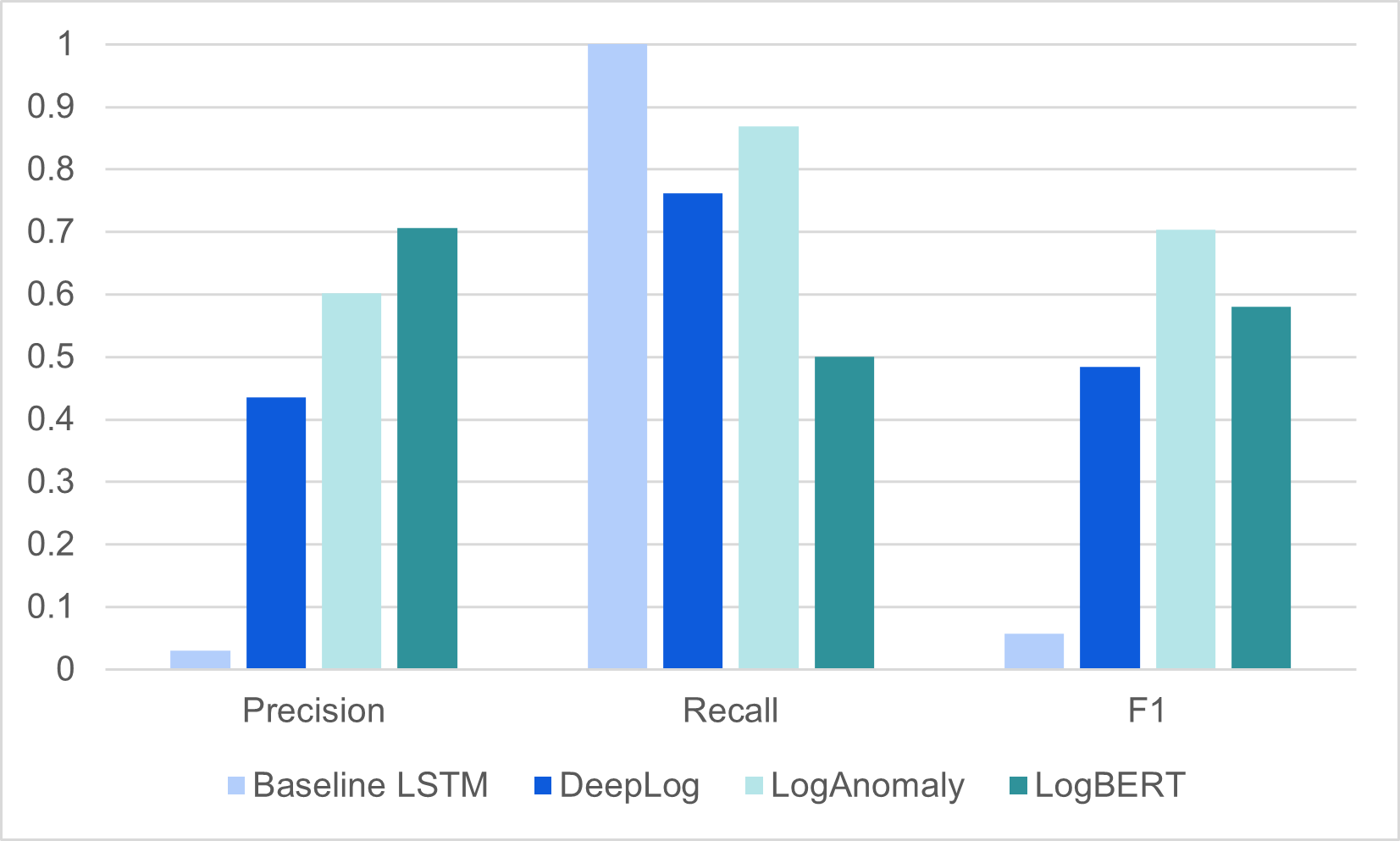}
    \caption{Model effectiveness on the industrial microservice dataset using random split. Each bar reports the corresponding metric's average over 5 folds.}
    \label{fig:aws_random}
\end{figure}

\begin{figure}[!tbp]
    \centering
    \includegraphics[width=0.45\textwidth]{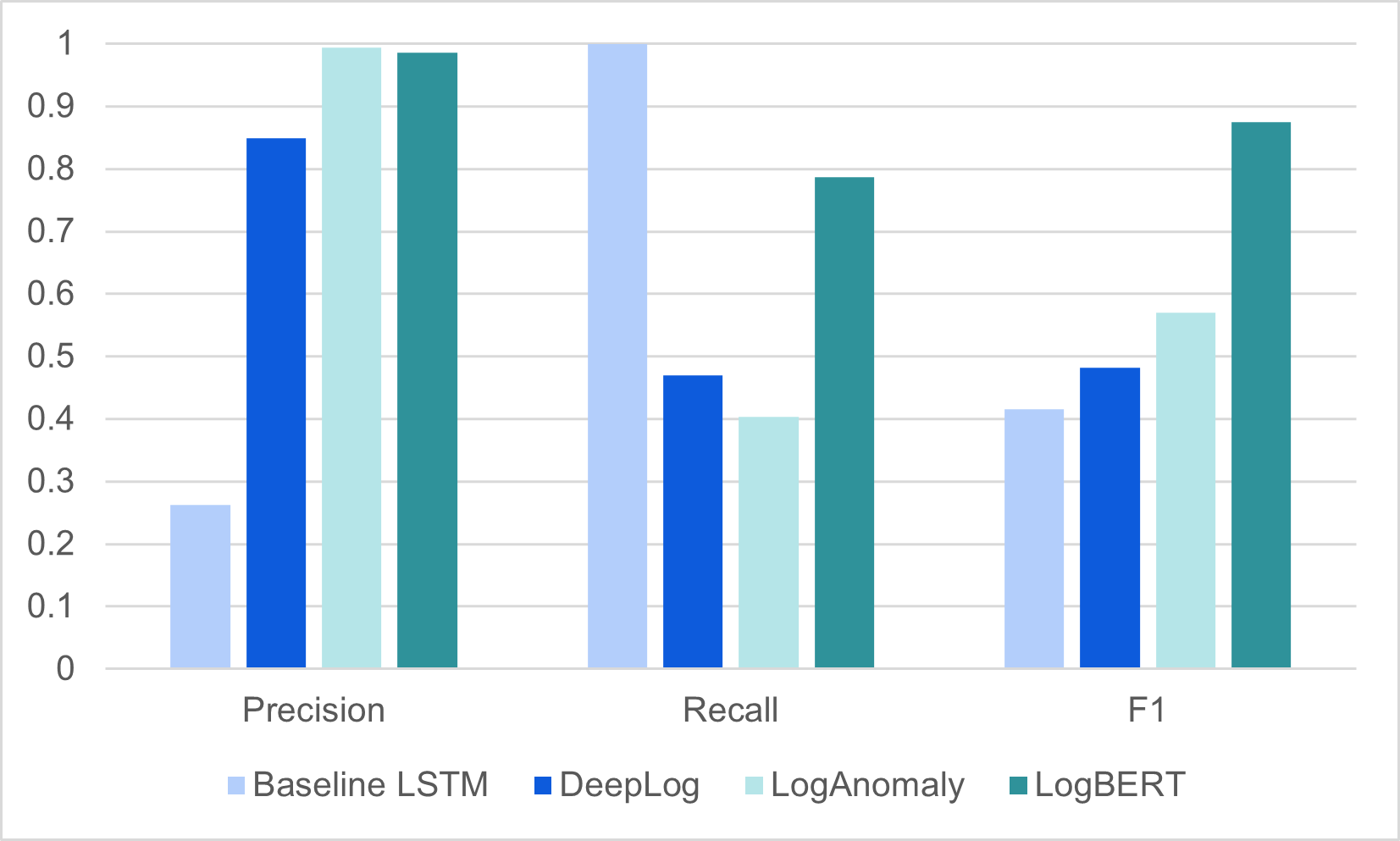}
    \caption{Model effectiveness on the HDFS dataset using random split. Each bar reports the corresponding metric's average over 5 folds.}
    \label{fig:hdfs_random}
\end{figure}

The models' effectiveness was measured using metrics outlined in the earlier section. The model performance on the industrial dataset is shown in figure \ref{fig:aws_random} and the HDFS dataset shown on figure \ref{fig:hdfs_random}.

For the industrial dataset, LogAnomaly has the best performance by far, with all other models significantly behind (the F1-score of LogAnomaly is 70.3\%, which is over 12\% higher than the F1-score of the other models). Even though the baseline method achieves perfect recall, this is undermined by its very low precision, and results with the lowest F1-score of all the models. 

On the contrary, when using the HDFS dataset, LogBERT outshines all other models with both F1-score (87.5\%) and recall (78.7\%). The F1-score of LogBERT is over 30\% higher than that of the other models, while the recall is over 31\% higher. While LogAnomaly is slightly higher in terms of precision, its low recall results in an overall lower effectiveness.

It should be noted that the baseline LSTM has the lowest performance out of all the models. This shows that a simple LSTM is not capable of inferring the patterns of log templates; additional logic is required.

The effectiveness of the models on the two datasets can be explained by the metrics and the nature of the datasets themselves. Based on table \ref{tab:dataset_properties}, it can be seen that while being smaller, the industrial dataset has a higher number of log templates, showing it's largely unstructured. Comparatively, the HDFS dataset has a much lower template count, even though it's significantly larger. LogAnomaly is shown capable of inferring relationships between logs with less structure, resulting in its high performance in the industrial dataset. LogBERT, on the other hand, needs a highly structured dataset to be of proper use. Even though the HDFS dataset is structured, only a very small amount (only 0.01\% of the entire dataset) was used as training, showing LogBERT can easily infer relationships even with a small training size, as long as the data itself is structured.

Comparing the mechanisms of the three state-of-the-art models, DeepLog and LogAnomaly both use LSTM networks and differ only in a few aspects. DeepLog makes use of parameter value vectors in addition to the log key anomaly detection LSTM, with the parameter values having their own models for each log key. LogAnomaly uses vectorization along with template approximation, and an added quantitative anomaly detection model. This vectorization step likely reduces noise in the logs, since minor updates to logging statements result in similar semantic vectors. It can also be hypothesized that template approximation helps with an unstructured dataset, as any templates that were not in the training set can be assigned a value. This could potentially be one of the reasons as to why LogAnomaly has a higher performance on the industrial dataset. On the other hand, having a large number of templates likely hinders the Masked Log Key Prediction (MLKP) task for LogBERT, since there can be more than one candidate for the missing [MASK] token. This likely results in mis-classifications during this step, reducing its effectiveness on loosely structured datasets.

The performance of DeepLog, LogAnomaly and LogBERT on the HDFS dataset is much lower than the metrics quoted in their original papers, showing the reduced training size severely impacted those models. The performance of LogBERT, however, is much closer to the metrics quoted in its origin paper, showing its ability to work even with a much smaller dataset.\\

\noindent\fbox{
    \parbox{.98\columnwidth}{
    \textbf{RQ1 Summary:}
The experiment results from three state-of-the-art models and the baseline LSTM model show that LogAnomaly works best with the industrial dataset, which is much less structured. LogBERT works best with the HDFS dataset, showing it requires a structured dataset to work effectively. The baseline LSTM has the weakest performance out of all the models.
    }
}\\

\subsubsection{RQ2: \RQTwo\ }

\paragraph{Design}

According to a recent study~\cite{hoang2022}, the data leak during random sampling makes the effectiveness of models seem higher than they actually are. Effectively, when doing random sampling, future logs are used as part of the training set, resulting in past logs being predicted using future logs. To explore this effect, we used a time-based split to train the models, while keeping the training size the same. This ensured that only past logs are used to predict future logs, which in turn helped us explore the realistic use-case of anomaly detection models: in practice, only past logs can be used to predict anomalies.

\begin{figure}[!tbp]
    \centering
    \includegraphics[width=0.45\textwidth]{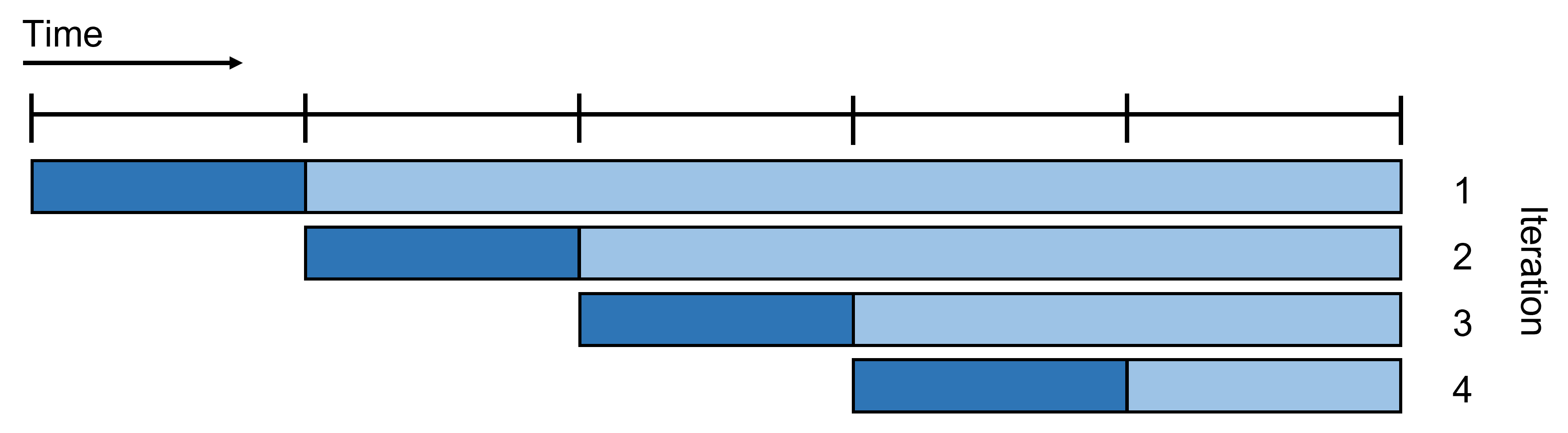}
    \caption{Splitting data based on time-series. Note that the dark-colored segments refer to the training set, while the light-colored segments are the test set.}
    \label{fig:time_series_split}
\end{figure}

Figure \ref{fig:time_series_split} shows the train and test sets created by using the time-series split. Note that by keeping the train size constant, only 4 folds could be created with this type of split; the last fold cannot be used for training, as there are no logs after it.

\paragraph{Results}

\begin{figure}[!bp]
    \centering
    \includegraphics[width=0.45\textwidth]{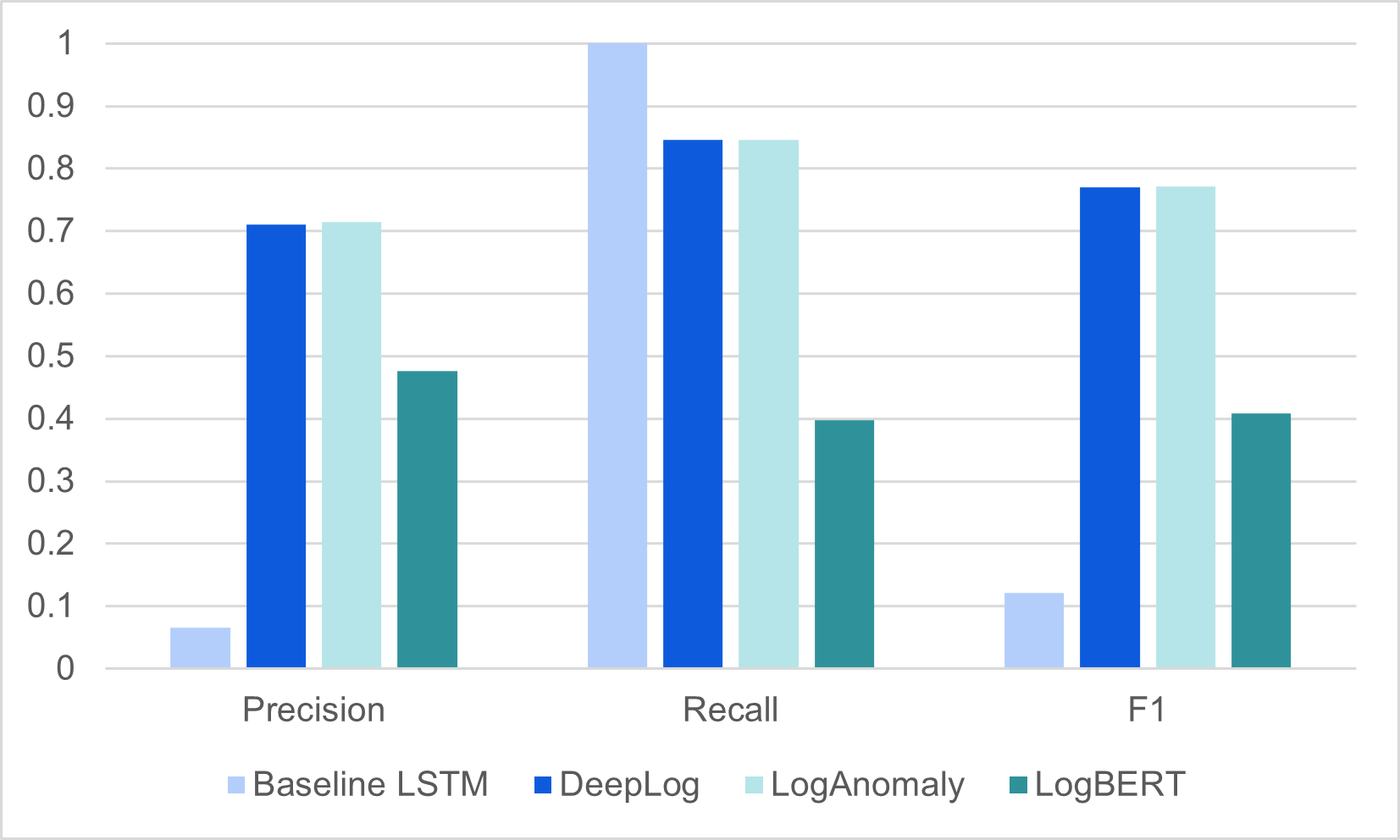}
    \caption{Model effectiveness on the industrial microservice dataset using time-based split. Each bar reports the corresponding metric's average over 4 folds.}
    \label{fig:aws_timebased}
\end{figure}

\begin{figure}[!tbp]
    \centering
    \includegraphics[width=0.45\textwidth]{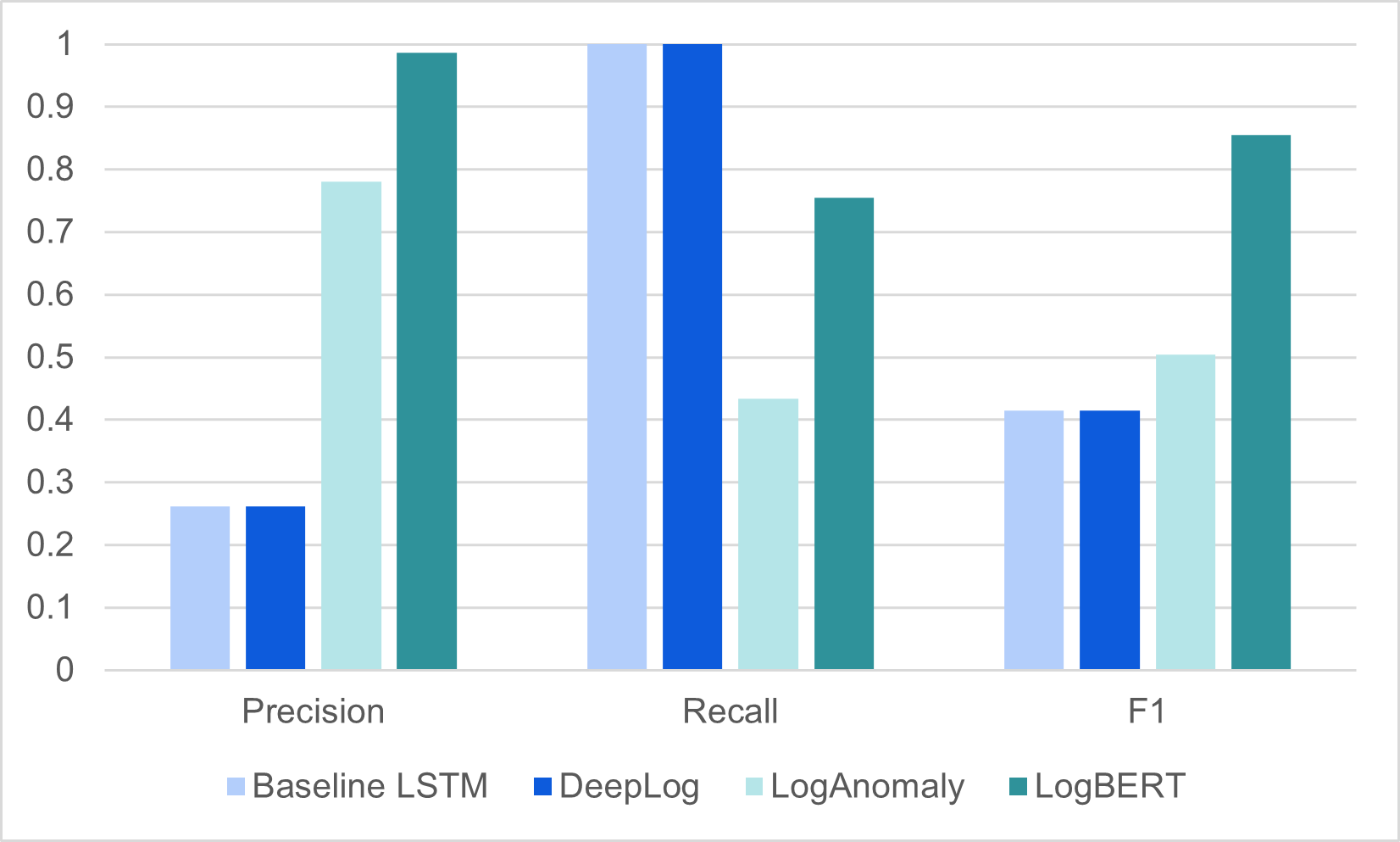}
    \caption{Model effectiveness on the HDFS dataset using time-based split. Each bar reports the corresponding metric's average over 4 folds.}
    \label{fig:hdfs_timebased}
\end{figure}

The model performance on the industrial dataset using a time-based split is shown in figure \ref{fig:aws_timebased} and the HDFS dataset is shown on figure \ref{fig:hdfs_timebased}.

Similar to the results from random split, LogAnomaly has the highest effectiveness with an F1-score of 77.2\%, but this time is closely followed by DeepLog with an F1-score of 77.0\%. They both have a similar recall as well, at approximately 84.6\%. LogBERT effectiveness has decreased significantly from earlier results, showing that the data leak plays an important role with increasing performance. Interestingly, DeepLog and LogAnomaly have increased in effectiveness. This is likely due to the fact that even with a time-based split, the amount of information conveyed through a single fold is sufficient for the model to be trained.

In the HDFS dataset, LogBERT outperforms the other models, similar to RQ1, with an F1-score of 85.5\%. DeepLog has the highest recall (at 100\%), but is undermined by its low precision, resulting in the lowest F1-score along with the baseline LSTM model (at 41.5\%). The effectiveness of all models have decreased from RQ1, showing with a large dataset, the data leak plays a bigger role in improving the model performance, compared to a smaller dataset. Similar to the earlier research question, we can see that LogBERT works best with a well-structured dataset, while LogAnomaly is more performant on the dataset with less structure.

In all of the above experiments, the baseline LSTM model has shown the weakest performance. Due to this, we have omitted it from the subsequent research questions, and focus only on the 3 state-of-the-art models.\\

\noindent\fbox{
    \parbox{.98\columnwidth}{
    \textbf{RQ2 Summary:}
Using a time-based split reduces the efficiency of the models for the most part, showing the data leak during random split plays a significant role in improving the model performance. LogBERT still shows a higher effectiveness with the HDFS dataset, while LogAnomaly works best with the industrial microservice dataset.
    }
}\\

\subsubsection{RQ3: \RQThree\ }

\paragraph{Design}

In the previous research questions, several existing anomaly detection models were evaluated on different datasets. In this section, a novel evaluation is performed on the model effectiveness when it comes to detecting specific types of errors. This aligns with the goal of selecting a suitable anomaly detection model for the application under study as well, since it gives us insights into how each model works with each type of errors.

In this question, we perform a qualitative analysis of the 3 state-of-the-art methods and the errors they detect in the industrial microservice dataset. By discussing with domain experts, we found 4 anomaly types, which can be classified into 2 categories, listed in table \ref{tab:aws_error_types}. Note that the anomaly types have been enumerated instead of being given descriptive names, for confidentiality reasons.

\begin{table}[!tbp]
    \caption{Anomaly types found in the industrial microservice dataset
\protect\label{tab:aws_error_types} }
    \begin{center}
      \begin{tabular}{|c|l|c|}
        \hline
        \textbf{Anomaly type} & \textbf{Anomaly category} & \textbf{Av. seq. length}\\
        \hline
        1 & Request error & 163.4\\
        2 & Request error & 170.25\\
        3 & Redis error & 109\\
        4 & Redis error & 7\\
        \hline
      \end{tabular}
    \end{center}
  \end{table}

A brief description of each of the anomaly categories is provided below:

\begin{itemize}
  \item \textbf{Request error}: Occurs when the request made to the server is invalid. This could be due to various reasons, such as a feature not being available in a region, invalid request payload etc.
  \item \textbf{Redis error}: Redis~\cite{redis} is an open-source data structure store, that's used as a cache for the servers. The errors of this type often stem from timeouts, which occur due to bandwidth limits, too many requests and high CPU/memory usages~\cite{redis_timeouts}.
\end{itemize}

To explore the effectiveness of each model in detecting the different types of anomalies, we trained the models using a 5-fold time-series split, and examined whether each error type was detected by the model. Since the model was trained using only normal logs, we used the entire anomalous sequences set for testing.

\paragraph{Results}

\begin{figure}[!tbp]
    \centering
    \includegraphics[width=0.45\textwidth]{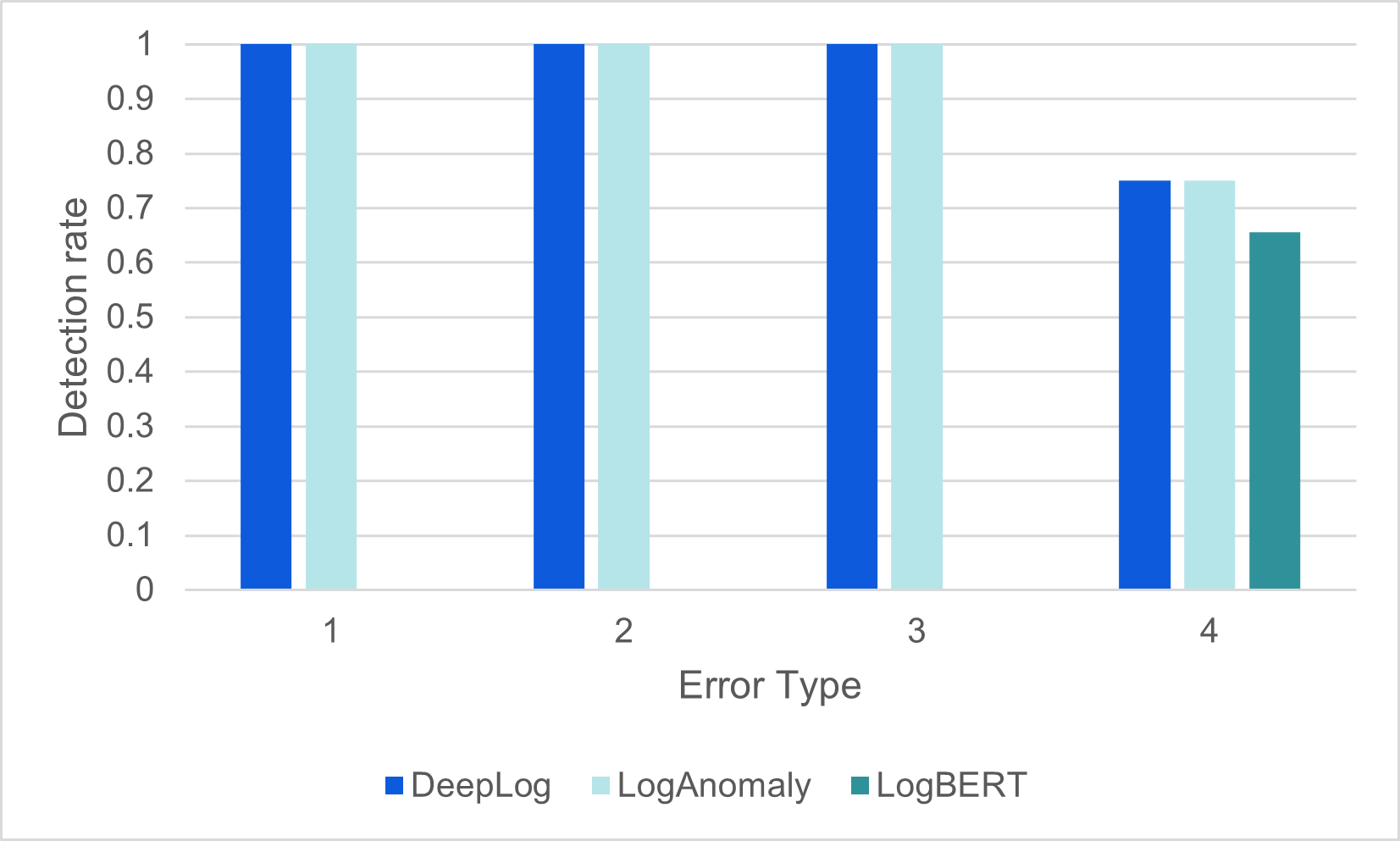}
    \caption{Model anomaly detection rates}
    \label{fig:error_detection_rate}
\end{figure}

Figure \ref{fig:error_detection_rate} shows the ability of each of the 3 models in detecting the 4 anomaly types (note that the detection rate for type 1, 2 and 3 anomalies for LogBERT is zero).

In the earlier research question, DeepLog and LogAnomaly had the best results with the industrial dataset time-series split, which is shown again by the results above. Both DeepLog and LogAnomaly are capable of fully detecting type 1, 2 and 3 anomalies, and has a comparatively higher performance on the type 4 anomalies as well. LogBERT, however, can only detect type 4 anomalies, but even this is poor compared to the other two models.

Comparing the log sequence length from Section \ref{tab:aws_error_types}, it seems LogBERT only partially works with very short sequences (type 4 has an average sequence length of 7, compared to other sequences which are over 100 in average length). Even though DeepLog has the same recall as LogAnomaly (at 84.6\%), it should be noted that its precision is slightly less than that of LogAnomaly, showing LogAnomaly as the overall best model in this scenario.\\

\noindent\fbox{
    \parbox{.98\columnwidth}{
    \textbf{RQ3 Summary:}
While LogBERT is capable of detecting short sequence anomalies, it does not work with longer sequences. DeepLog and LogAnomaly both work with long sequence anomalies, and compared to LogBERT has a better performance with short sequence anomalies as well.
    }
}\\

\subsubsection{RQ4: \RQFour\ }

\paragraph{Design}

One of the driving factors in our research has been the lack of rich training data during early development stages for most systems under test. Due to this, we have opted for small training sizes, with a split type comparable to that of papers done on open-source data. In this research question, we explore if it is possible to improve the effectiveness of the models by increasing the size of the training set. This would give us an insight as to which of the parameters play a bigger role in model performance: data size or data uniformity.

The experiment was performed using a time-series split, with increasing train sizes from 20\% up to 80\%, in 20\% increments.

\paragraph{Results}

\begin{figure}[!tbp]
    \centering
    \includegraphics[width=0.45\textwidth]{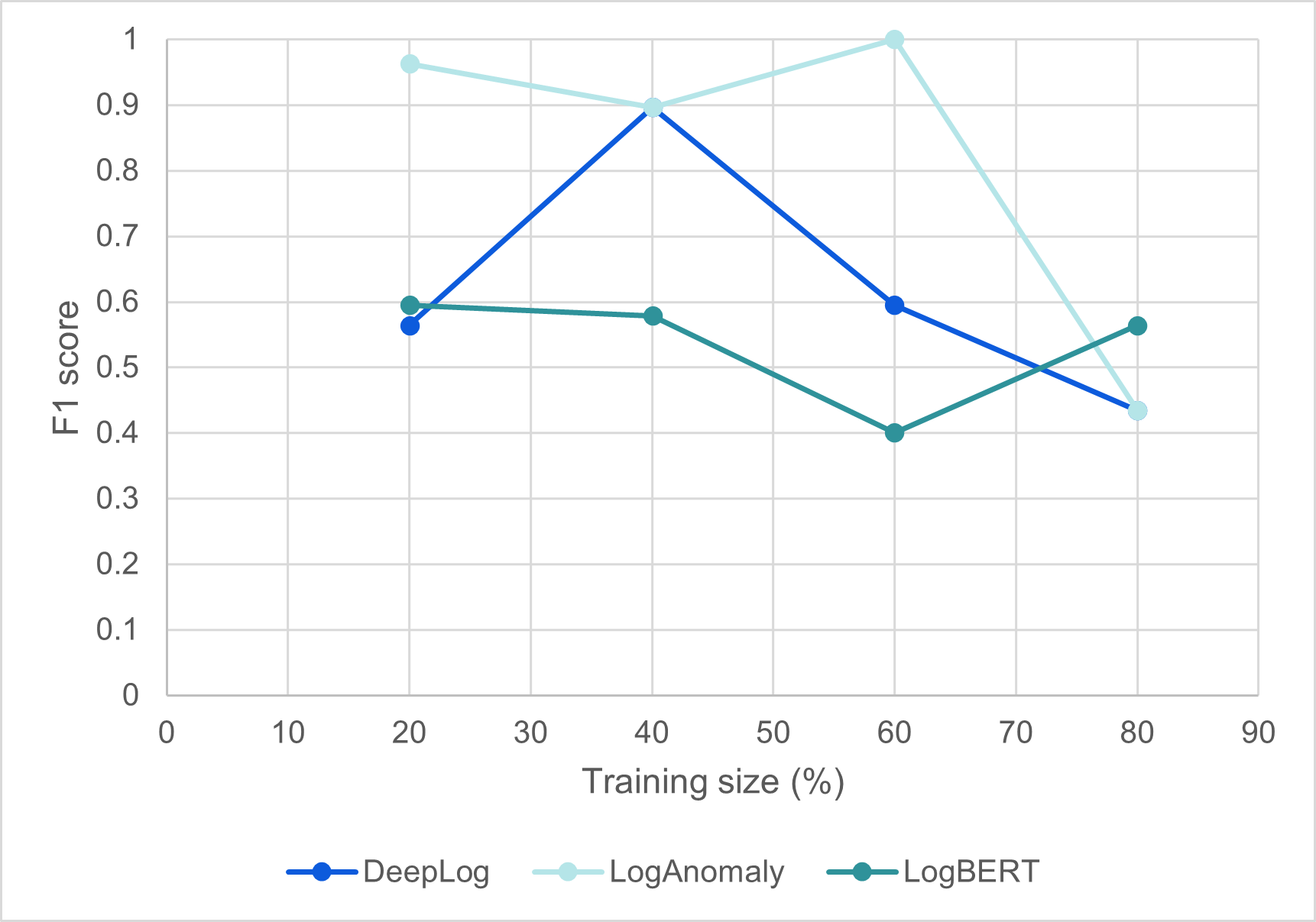}
    \caption{F1-score dependency on training size}
    \label{fig:rq4_chart}
\end{figure}

The results are shown in figure \ref{fig:rq4_chart}. Here, we can see that LogBERT has the poorest performance, with an F1-score below 60\% that does not increase with an increasing training size. This further shows that LogBERT is not efficient at inferring patterns in a loosely-structured dataset. DeepLog performance does get better with an increasing training size, peaking at 40\% and then dropping, likely due to overfitting. LogAnomaly also gets better with increasing train size, and achieves a perfect score of 100\% F1-score at a training size of 60\%. This too, then drops at 80\% training size, also possibly due to overfitting.

Overall, LogAnomaly works best when the training size is limited, and increases comparatively better when the training size is increased. Having a better structured dataset may help LogBERT achieve a better score, which we have not tested in this experiment.\\

\noindent\fbox{
    \parbox{.98\columnwidth}{
    \textbf{RQ4 Summary:}
Increasing the training size using the industrial microservice data helps improve performance of DeepLog and LogAnomaly, which decreases after reaching a peak, showing effects of overfitting. LogBERT does not get much better with an increased training size, showing it does not work effectively with a loosely-structured dataset.
    }
}\\

\subsection{Threats to Validity}

\textbf{Internal validity.} Internal validity refers to unforeseen factors that may influence the outcome of the experiments. One such aspect are the hyper-parameters used for the models under test, as different values for the hyper-parameters can affect the performance of the models. For consistency, we have used the default values for each of the models in this test, barring several configuration values that needed to be set to allow for shorter sequences of logs from the industrial microservice. The parameters for the HDFS experiments, however, have not been changed from their default values. This should allow for a direct comparison with earlier papers using the open-source datasets. Further hyper-parameter turning of the models with the industrial dataset, however, may improve the performance of those models.

\textbf{External validity.} The external threats to validity include ability to generalize the results of the study. There are two such major external threats to validity in this study. First is the limited dataset, as the experiments were conducted on a single microservice. Application of these models on further services would increase the exposure of the research. Second is the limited amount of faults in the industrial dataset. As is the case with most stable industrial applications, the amount of anomalous sequences is very limited, compared to the amount of normal sequences. We believe, however, that it makes this problem represent real-world problem, and improves the applicability of this experiment on other stable applications.

\section{Conclusion and Future Work}
In this paper, we have conducted an application of several anomaly detection models on an industrial dataset, with a real-world limitation on the dataset size and log data uniformity. Results suggest that the LogAnomaly model works best on less structured datasets, such as our industrial dataset. A qualitative analysis on the anomaly types by the experts showed that LogAnomaly and DeepLog are both effective at detecting different types of anomalies with short and long sequences, while LogBERT struggles with even short sequence ones. Exploring the effect of the training size shows that LogAnomaly and DeepLog do get better results with a bigger training set, but over-sized training sets result in the model over-fitting and a reduced effectiveness. In conclusion, LogAnomaly was identified as the overall best performing model in our case study.

Future work on this paper can be pursued on several avenues. First, the model hyper-parameters can be tuned to work better with the industrial dataset, possibly improving their performance. Second, more models, including classical models such as SVM and LogClustering can be applied on the dataset, which can provide another set of baselines to compare against. State-of-the-art models such as LogRobust~\cite{zhang2019}, which have built-in vectorization, would also be a good candidate for dealing with loosely-structured logs.
Another possible avenue of future work is using vectorization, by using a natural language model such as a Generative Pre-trained Transformer (GPT). With this, the template mining step would become optional, and the log events can be directly vectorized. This would help with continuously evolving log messages, as well as noise within the logs.
Another possible extension is to evaluate the models on more industrial datasets, which may come with their own limitations. This would shed more light into the applicability of those models on more real-world systems.

\bibliographystyle{IEEEtran}
\bibliography{references}

\end{document}